\documentclass[11pt,a4paper]{article}
\usepackage[hyperref]{emnlp-ijcnlp-2019}
\usepackage{times}
\usepackage{latexsym}
\usepackage{CJKutf8}
\usepackage{multirow}
\usepackage{arydshln}
\usepackage{booktabs}
\usepackage{amssymb}
\usepackage{color,soul}
\usepackage{enumitem}

\usepackage{url}

\aclfinalcopy 

\newcommand{\Yes}{\checkmark}
\newcommand{\No}{$\times$}
\newcommand{\xpaws}{PAWS-X}

\title{\xpaws: A Cross-lingual Adversarial Dataset for Paraphrase Identification}

\author{Yinfei Yang\thanks{\hspace{0.5em}equal contribution},~~ Yuan Zhang\footnotemark[1],~~ Chris Tar \and Jason Baldridge \\
Google Research, Mountain View, 94043 \\
  {\tt \{yinfeiy, zhangyua, ctar, jasonbaldridge\}@google.com}
}

\date{}

\begin{document}
\maketitle

\begin{abstract}
Most existing work on adversarial data generation focuses on English. For example, PAWS (Paraphrase Adversaries from Word Scrambling) \cite{paws2019naacl} consists of challenging English paraphrase identification pairs from Wikipedia and Quora. We remedy this gap with \xpaws, a new dataset of 23,659 \emph{human} translated PAWS evaluation pairs in six typologically distinct languages: French, Spanish, German, Chinese, Japanese, and Korean.  We provide baseline numbers for three models with different capacity to capture non-local context and sentence structure, and using different multilingual training and evaluation regimes. Multilingual BERT \cite{devlin2018bert} fine-tuned on PAWS English plus machine-translated data performs the best, with a range of 83.1-90.8 accuracy across the non-English languages and an average accuracy gain of 23\% over the next best model. \xpaws\ shows the effectiveness of deep, multilingual pre-training while also leaving considerable headroom as a new challenge to drive multilingual research that better captures structure and contextual information.
\end{abstract}

\section{Introduction}

Adversarial examples have effectively highlighted the deficiencies of state-of-the-art models for many natural language processing tasks, e.g. question answering \cite{JiaL17,chen2018attacking,ribeiro2018semantically}, textual entailment \cite{zhao2018generating,glockner2018breaking}, and text classification \cite{alzantot2018generating,IyyerWGZ18}. 
\newcite{paws2019naacl} introduce PAWS, which has adversarial paraphrase identification pairs with high lexical overlap, like \textit{flights from New York to Florida} and \textit{flights from Florida to New York}.
Such pairs stress the importance of modeling sentence structure and context because they have high word overlap ratio but different semantic meaning.
In addition to revealing failures of state-of-the-art models, research on adversarial examples has generally shown that augmenting training data with good adversarial examples can boost performance for some models---providing greater clarity to the modeling landscape as well providing new headroom for further improvements. 

Most previous work focuses only on English despite the fact that the problems highlighted by adversarial examples are shared by other languages. Existing multilingual datasets for paraphrase identification, e.g. Multi30k \cite{desmond2016multi30k} and Opusparcus \cite{creutz-2018-open}, lack challenging examples like PAWS. 
The lack of high-quality adversarial examples in other languages makes it difficult to benchmark model improvements. We bridge this gap by creating Cross-lingual PAWS (\xpaws), an extension of the Wikipedia portion of the PAWS evaluation and test examples to six languages: Spanish, French, German, Chinese, Japanese, and Korean. This new corpus consists of 23,659 \textit{human} translated example pairs with paraphrase judgments in each target language. Like previous work on multilingual corpus creation \cite{conneau-etal-2018-xnli}, we machine translate the original PAWS English training set (49,401 pairs).
Note that all translated pairs still have high word overlap and they inherit semantic similarity labels from the original PAWS examples; thus, the resulting dataset preserves the ability of probing structure and context sensitivity for models. 
We also machine translate the evaluation pairs of each language into English to establish the baseline performance of a translate-then-predict strategy. The \xpaws\ dataset, including both the new human translated pairs and the machine translated examples, is available for download at \href{https://github.com/google-research-datasets/paws}{https://github.com/google-research}\href{https://github.com/google-research-datasets/paws}{-datasets/paws}.

\begin{table*}[!htb]
    \small
    \centering
    \begin{tabular}{c|l}
        \hline
        {\bf Language} & \multicolumn{1}{c}{\bf Text} \\ \hline
        \multicolumn{2}{c}{\rule{0pt}{8pt} \textit{Original Pair (id: 000005309\_9438, label: not-paraphrasing) }  \rule{0pt}{8pt}} \\ \hline
        \multirow{2}{*}{\bf en} 
            & However, in order to defeat Slovak, Derek must become a vampire attacker. \\
            & However, in order to become Slovak, Derek must defeat a vampire assassin.	\\ \hline
        \multicolumn{2}{c}{\rule{0pt}{8pt} \textit{Human Translated Pairs} \rule{0pt}{8pt}} \\ \hline
        \multirow{2}{*}{\bf fr} 
            &  Toutefois, pour battre Slovak, Derek doit devenir un vampire attaquant. \\
            &  Cependant, pour devenir Slovak, Derek doit vaincre un vampire assassin. \\ \cdashline{1-2}
        \multirow{2}{*}{\bf es} 
            &  Sin embargo, para derrotar a Slovak, Derek debe convertirse en un atacante vampiro. \\
            &  Sin embargo, para poder convertirse en Slovak, Derek debe derrotar a un asesino de vampiros. \\ \cdashline{1-2}
        \multirow{2}{*}{\bf de} 
            &  Um Slovak zu besiegen, muss Derek jedoch zum Vampirjäger werden. \\
            &  Um jedoch Slowake zu werden, muss Derek einen Vampirjäger besiegen. \\ \cdashline{1-2}
        \multirow{2}{*}{\bf zh} 
            & \begin{CJK}{UTF8}{gbsn}但为击败斯洛伐克，德里克必须成为吸血鬼攻击者。\end{CJK}\\
            & \begin{CJK}{UTF8}{gbsn}然而，为了成为斯洛伐克人，德里克必须击败吸血鬼刺客。 \end{CJK} \\ \cdashline{1-2}
        \multirow{2}{*}{\bf ja} 
            & \begin{CJK}{UTF8}{min}ただし、スロバークを倒すためには、デレクは吸血鬼アタッカーになる必要があります。 \end{CJK} \\
            & \begin{CJK}{UTF8}{min}しかし、デレクがスロバックになるには、バンバイア・アサシンを倒さなければならない。 \end{CJK} \\ \cdashline{1-2}
        \multirow{2}{*}{\bf ko} 
            & \begin{CJK}{UTF8}{mj}하지만 Slovak이 되기 위해 Derek은 반드시 뱀파이어 암살자를 물리쳐야만 합니다. \end{CJK} \\
            & \begin{CJK}{UTF8}{mj}하지만 Slovak을 물리치기 위해 Derek은 뱀파이어 사냥꾼이 되어야만 했습니다. \end{CJK} \\ \hline
    \end{tabular}
    \caption{Examples of human translated pairs for each of the six languages.}
    \label{tab:data_examples}
\end{table*}

Our experiments show that \xpaws\ effectively measures the multilingual adaptability of models and how well they capture context and word order. The state-of-the-art multilingual BERT model \cite{devlin2018bert} obtains a 32\% (absolute) accuracy improvement over a bag-of-words model. We also show that machine translation helps and works better than a zero-shot strategy. We find that performance on German, French, Spanish is overall better than Chinese, Japanese and Korean.

\section{\xpaws\ Corpus}
\label{sec:corpus}

The core of our corpus creation procedure is to translate the Wikipedia portion of the original PAWS corpus from English (en) to six languages: French (fr), Spanish (es), German (de), Chinese (zh), Japanese (ja), and Korean (ko). To this end, we hire human translators to translate the development and test sets, and use a neural machine translation (NMT) service\footnote{\url{https://cloud.google.com/translate/}} to translate the training set.

We choose translation instead of repeating the PAWS data generation approach \cite{paws2019naacl} to other languages. This has at least three advantages. First, human translation does not require high-quality multilingual part-of-speech taggers or named entity recognizers, which play a key role in the data generation process used in \citet{paws2019naacl}. Second, human translators are trained to produce the target sentence while preserving meaning, thereby ensuring high data quality. Third, the resulting data can provide a new testbed for cross-lingual transfer techniques because examples in all languages are translated from the same sources. For example, \xpaws\ could be used to evaluate whether a German or French sentence is a paraphrase of a Chinese or Japanese one.

\begin{table}
    \small
    \centering
    \begin{tabular}{c|cccccc}
           & {\bf fr} & {\bf es} & {\bf de} & {\bf zh} & {\bf ja} & {\bf ko} \\ \hline
        dev     & 1,992 & 1,962 & 1,932 & 1,984 & 1,980 & 1,965 \\  
        test    & 1,985 & 1,999 & 1,967 & 1,975 & 1,946 & 1,972 \\  
    \end{tabular}
     \caption{Examples translated per language.}
     \label{tab:final_data}
\end{table}

\paragraph{Translating Evaluation Sets} We obtain human translations on a random sample of 4,000 sentence pairs from the PAWS development set for each of the six languages (48,000 translations). The manual translation is performed by 10-20 in-house professionals that are native speakers of each language. A randomly sampled subset is presented and validated by a second worker. The final delivery is guaranteed to have less than 5\% word level error rate. The sampled 4,000 pairs are split into new development and test sets, 2,000 pairs for each.

Due to time and cost constraints, we could not translate all 16,000 examples in both of original PAWS development and test set. Each sentence in a pair is presented independently so that translation is not affected by context. In our initial studies we noticed that sometimes it was difficult to translate an entity mention. We therefore ask translators to translate entity mentions, but different translators may have different preferences according to their background knowledge. Table \ref{tab:data_examples} gives example translated pairs in each language.

\paragraph{Resulting Corpus} Some sentences could not be be translated. Table \ref{tab:final_data} shows the final counts translated to each language. Most of the untranslated sentences were due to incompleteness or ambiguities, such as \textit{It said that Easipower was}, and \textit{Park Green took over No}. These sentences are likely from the adversarial generation process when creating PAWS. On average less than 2\% of the pairs are not translated, and we simply exclude them. 

The authors further verified translation quality for a random sample of ten pairs in each language. \xpaws\ includes 23,459 human-translated pairs, including 11,815 and 11,844 pairs in development and test, respectively. Finally, original PAWS labels (paraphrase or not paraphrase) are mapped to the translations. Positive pairs account for 44.0\% of development sets and 45.4\% of test respectively--close to the PAWS label distribution.

Translation brings new challenges to the paraphrasing identification task. An entity can be translated differently, such as \textit{Slovak} and \textit{Slowake} (Table \ref{tab:data_examples}) and models need to capture that these refer to the same entity. In a more challenging example, \textit{Four Rivers}, \textit{Audubon} and \textit{Shawnee Trails} are translated in just one of the sentences:

\begin{small}
\begin{itemize}[leftmargin=.36in]
    \item[{\bf en} \textit{s1}] From the merger of the Four Rivers Council and the Audubon Council, the Shawnee Trails Council was born.
    \vspace{-0.1in}
    \item[\textit{s2}] Shawnee Trails Council was formed from the merger of the Four Rivers Council and the Audubon Council.
    \item[{\bf zh} \textit{s1}] \begin{CJK}{UTF8}{gbsn}Four Rivers 委员会与 Audubon 委员会合并后，Shawnee Trails 委员会得以问世. \end{CJK} 
    \vspace{-0.1in}
    \item[\textit{s2}] \begin{CJK}{UTF8}{gbsn}\underline{肖尼小径(Shawnee Trails)} 委员会由合并 \underline{四河 (Four Rivers)} 委员会和 \underline{奥杜邦 (Audubon)} 委员会成立. \end{CJK}
\end{itemize}
\end{small}

\noindent
In the zh-s2 example, the parentheses give English glosses of Chinese entity mentions.
\section{Evaluated Methods}

The goal of \xpaws\ is to probe models' ability to capture structure and context in a multilingual setting. We consider three models with varied complexity and expressiveness. The first baseline is a simple bag-of-words (\textbf{BOW}) encoder with cosine similarity. It uses unigram to bigram token encoding as input features and takes a cosine value above 0.5 as a paraphrase. The second model is \textbf{ESIM}, Enhanced Sequential Inference Model \cite{ChenZLWJI17}. Following \citet{paws2019naacl}, ESIM encodes each sentence using a BiLSTM, and passes the concatenation of encodings through a feed-forward layer for classification. The additional layers allow ESIM to capture more complex sentence interaction than cosine similarity.  Third, we evaluate \textbf{BERT}, Bidirectional Encoder Representations from Transformers \cite{devlin2018bert}, which recently achieved state-of-the-art results on eleven natural language processing tasks. 

We evaluate all models with two strategies \cite{conneau-etal-2018-xnli}: (1) {\bf Translate Train}: the English training data is machine-translated into each target language to provide data to train each model and (2) {\bf Translate Test}: train a model using the English training data, and machine-translate all test examples to English for evaluation.

Multilingual BERT is a single model trained on 104 languages, which enables experiments with cross-lingual training regimes. (1) {\bf Zero Shot}: the model is trained on the PAWS English training data, and then directly evaluated on all others. Machine translation is {\bf not} involved in this strategy. (2) {\bf Merged}: train a multilingual model on all languages, including the original English pairs and machine-translated data in all other languages.

Table \ref{tb:models} summarizes the models with respect to whether they represent non-local contexts or support cross-sentential word interaction, plus which strategies are evaluated for each model.

\begin{table}[t]
    \centering
    \small
    \begin{tabular}{l|ccc}
      & BOW & ESIM & BERT  \\
    \hline
    Non-local context & \No & \Yes & \Yes \\
    Word interaction & \No & \No & \Yes \\
    \hline
    Translate Train & \Yes & \Yes & \Yes \\
    Translate Test  & \Yes & \Yes & \Yes \\
    Zero Shot       & \No  & \No  & \Yes \\
    Merged          & \No  & \No  & \Yes \\
    \end{tabular}
    \caption{Complexity of each evaluated model and the training/evaluation strategies being tested.}
    \label{tb:models}
\end{table}

{
\begin{table*}[ht!]
    \small
    \begin{center}
        \begin{tabular}{@{~~~}l|c@{~~~~} c@{~~~~} c@{~~~~} c@{~~~~} c@{~~~~} c@{~~~~} c@{~~~~} |c@{~~~~} c@{~~~~} c@{~~~~} c@{~~~~} c@{~~~~} c@{~~~~} c@{~~~~}}
        \multirow{2}{*}{\textbf{Method}} & \multicolumn{7}{c|}{\bf Accuracy} & \multicolumn{7}{c}{\bf AUC-PR} \\
        & \bf{en} & \bf{fr} & \bf{es} & \bf{de} & \bf{zh} & \bf{ja} & \bf{ko}  & \bf{en} & \bf{fr} & \bf{es} & \bf{de} & \bf{zh} & \bf{ja} & \bf{ko} \\
        \hline
        \rule{0pt}{8pt}
        \textbf{BOW} & & & & & & & & & & & & & & \\
        ~~~Translate Train     & 55.8 & 51.7 & 47.9 & 50.2 & 54.5 & 55.1 & 56.7 & 41.1 & 48.9 & 46.8 & 46.4 & 50.0 & 48.7 & 49.3 \\ 
        ~~~Translate Test      &  --  & 54.9 & 54.7 & 55.2 & 55.3 & 55.9 & 55.2 & --   & 46.3 & 45.5 & 45.8 & 50.9 & 46.8 & 48.5 \\
        \rule{0pt}{8pt}
        \textbf{ESIM} & & & & & & & & & & & & & & \\
        ~~~Translate Train   & 67.2 & 66.2 & 66.0 & 63.7 & 60.3 & 59.6 & 54.2 & 69.6 & 67.0 & 64.2 & 59.2 & 58.2 & 56.3 & 50.5  \\ 
        ~~~Translate Test    &  --  & 66.2 & 66.3 & 66.0 & 62.0 & 62.3 & 60.6 & --   & 68.4 & 69.5 & 68.2 & 62.3 & 61.8 & 60.3 \\
        \rule{0pt}{8pt}
        \textbf{BERT} & & & & & & & & & & & & & & \\
        ~~~Translate Train     & 93.5 & 89.3 & 89.0 & 85.3 & 82.3 & 79.2 & 79.9 & \bf{97.1} & 93.6 & 92.4 & 92.0 & 87.4 & 81.4 & 82.4  \\ 
        ~~~Translate Test    &  --  & 88.7 & 89.3 & 88.4 & 79.3 & 75.3 & 72.6 &  --  & 93.8 & 93.1 & 92.9 & 85.1 & 80.9 & 80.1  \\
        ~~~Zero shot           &  --  & 85.2 & 86.0 & 82.2 & 75.8 & 70.5 & 71.7 &  --  & 91.0 & 90.5 & 89.4 & 79.6 & 72.7 & 75.5  \\
        ~~~Merged              & \bf{93.8} & \bf{90.8} & \bf{90.7} & \bf{89.2} & \bf{85.4} & \bf{83.1} & \bf{83.9} & 96.5 & \bf{94.0} & \bf{92.9} & \bf{92.9} & \bf{88.9} & \bf{86.0} & \bf{86.3}\\
        \end{tabular}
    \end{center}
    \caption{Accuracy (\%) and AUC-PR (\%) of each approach. Best numbers in each column are marked in bold.}
    \label{tab:results_all}
    \vspace{-0.15in}
\end{table*}
}

{
\begin{table}[t]
    \small
    \begin{center}
        \begin{tabular}{l  l |  c c}
             \multicolumn{2}{l|}{\multirow{2}{*}{{\bf Method}}} & \multicolumn{2}{c}{\bf Averaged} \\
             \multicolumn{2}{l|}{} & Accuracy & AUC-PR \\ \hline
            \rule{0pt}{8pt}
             \multirow{2}{*}{\bf BOW}   & Translate Train   & 52.7 & 48.4 \\
                                        & Translate Test  & 55.2 & 47.3 \\ \cdashline{1-4}
            \rule{0pt}{8pt}
             \multirow{2}{*}{\bf ESIM}  & Translate Train   & 61.7 & 59.2 \\
                                        & Translate Test  & 63.9 & 65.1 \\  \cdashline{1-4}
            \rule{0pt}{8pt}
             \multirow{4}{*}{\bf BERT}  & Translate Train   & 84.2 & 88.2 \\
                                        & Translate Test  & 82.3 & 87.6 \\ 
                                        & Zero Shot         & 78.6 & 83.1 \\ 
                                        & Merged      & \textbf{87.2} & \textbf{90.2} \\
        \end{tabular}
    \end{center}
    \caption{Average Accuracy (\%) and AUC-PR (\%) over the six languages.}
    \label{tab:results_avg}
    \vspace{-0.1in}
\end{table}
}

\section{Experiments and Results}
We use the latest public multilingual BERT base model with 12 layers\footnote{\url{http://goo.gl/language/bert}} and apply the default fine-tuning strategy with batch size 32 and learning rate 1e-5. For BOW and ESIM, we use our own implementations and 300 dimensional multilingual word embeddings from fastText.\footnote{\url{https://fasttext.cc/}} We allow fine-tuning word embeddings during training, which gives better empirical performance.


We use two metrics: classification accuracy and area-under-curve scores of precision-recall curves (AUC-PR). For BERT, probability scores for the positive class is used to compute AUC-PR. For BOW and ESIM a cosine threshold of 0.5  is used to compute accuracy. In all experiments, the best model checkpoint is chosen based on accuracy on development sets and report results on testing sets.

\paragraph{Results} Table \ref{tab:results_all} shows the performance of all methods and languages. Table \ref{tab:results_avg} summarizes the average results for the six non-English languages.

\textbf{Model Comparisons:} On both Translate Train and Translate Test, BERT consistently outperforms both BOW and ESIM by a substantial margin ($>$15\% absolute accuracy gains) across all seven languages. BERT Translate Train achieves an average 20\% accuracy gain. This result demonstrates that \xpaws\ effectively measures models’ sensitivity to word order and syntactic structure.

\textbf{Training/Evaluation Strategies:} As Table \ref{tab:results_all} and \ref{tab:results_avg} show, the Zero Shot strategy yields the lowest performance compared to other strategies on BERT. This is evidence that machine-translated data helps in the multilingual scenario. Indeed, when training on machine-translated examples in all languages (Merged), the model achieves the best performance, with 8.6\% accuracy and 7.1\% AUC-PR average gains over Zero Shot. 

BERT and ESIM show different performance patterns on Translate Train and Translate Test. Translate Test appears to give consistently better performance then Translate Train on ESIM, but not on BERT. This may be because multilingual BERT is pre-trained on over one hundred languages; hence BERT provides better initialization for non-English languages than ESIM (which relies on fastText embeddings). The gap between training on English and on other languages is therefore smaller on BERT than on ESIM, which makes Translate Train work better on BERT.

\textbf{Language Difference:} Across all models and approaches, performance on Indo-European languages (German, French, Spanish) is consistently better than CJK (Chinese, Japanese, Korean). The performance difference is particularly noticeable on Zero Shot. This can be explained from two perspectives. First, the MT system we used works better on Indo-European languages than on CJK. Second, the CJK family is more typologically and syntactically different from English. For example, in table \ref{tab:data_examples},  \textit{Slowake} in German is much closer to the original term \textit{Slovak} in English, compared with its Chinese translation \begin{CJK}{UTF8}{gbsn}\textit{斯洛伐克}\end{CJK}.
This at least partly explains why performance on CJK is particularly poor in Zero Shot.

\begin{table}[!tb]
    \small
    \begin{center}
        \begin{tabular}{l  |  r r r r r}
        & 0 & 1-2 & 3-4 & 5-6 & 7 \\
        \hline
        \# & 32 & 52 & 140 & 542 & 1234 \\
        \% & 1.6 & 2.6 & 7.0 & 27.1 & 61.7
        \end{tabular}
    \end{center}
    \caption{The count of examples by number of languages (of 7) that agree with the gold label in test set.}
    \label{tab:agreement_with_gold}
    \vspace{-0.05in}
\end{table}

\textbf{Error Analysis:} To gauge the difficulty of each example for the best model (BERT-merged), Table \ref{tab:agreement_with_gold} shows the count of examples based on how many languages for the same pair are assigned the correct label in test set. The majority of the examples are easy, with 61.7\% correct in all languages. Of the 32 examples that failed in all languages, most are hard or highly ambiguous. Some have incorrect gold labels or were generated incorrectly in the original PAWS data.

The following is a sample of these.

\begin{itemize}[leftmargin=.2in]
\small
    \item[a1] On July 29, 1791, Sarah married Lea Thomas Wright Hill (1765--1842) at St. Martin's Church in Birmingham and had 8 children. 
    \vspace{-0.05in}
	\item[a2] Thomas Wright Hill married Sarah Lea (1765--1842) on 29 July 1791 at St Martin's Church, Birmingham and had 8 children. \textbf{match}
    \item[b1] He established himself eventually in the northwest of Italy, apparently supported by Guy, where he probably comes ``title''.
    \vspace{-0.05in}
	\item[b2] He eventually established himself in northwestern Italy, apparently supported by Guy, where he probably received the title of ``comes''. \textbf{not\_match}
    \vspace{-0.05in}
\end{itemize}

We also considered examples that are correctly predicted in just half of the languages. Some of these failed because of translation noise, e.g. inconsistent entity translations (as shown in \S\ref{sec:corpus}). 

\section{Conclusion}

We introduce \xpaws, a challenging paraphrase identification dataset with 23,659 \emph{human} translated evaluation pairs in six languages. Our experimental results showed that \xpaws\  effectively measures sensitivity of models to word order and the efficacy of cross-lingual learning approaches. It also leaves considerable headroom as a new challenging benchmark to drive multilingual research on the problem of paraphrase identification.

\section*{Acknowledgments}
We would like to thank our anonymous reviewers and the Google AI Language team, especially Luheng He, for the insightful comments that contributed to this paper. Many thanks also to the translate team, especially Mengmeng Niu, for the help with the annotations.

\bibliography{emnlp-ijcnlp-2019}

\begin{thebibliography}{13}
\expandafter\ifx\csname natexlab\endcsname\relax\def\natexlab#1{#1}\fi

\bibitem[{Alzantot et~al.(2018)Alzantot, Sharma, Elgohary, Ho, Srivastava, and
  Chang}]{alzantot2018generating}
Moustafa Alzantot, Yash Sharma, Ahmed Elgohary, Bo-Jhang Ho, Mani Srivastava,
  and Kai-Wei Chang. 2018.
\newblock {Generating natural
  language adversarial examples}.
\newblock In \emph{Proceedings of the 2018 Conference on Empirical Methods in
  Natural Language Processing}, pages 2890--2896, Brussels, Belgium.
  Association for Computational Linguistics.

\bibitem[{Chen et~al.(2018)Chen, Zhang, Chen, Yi, and
  Hsieh}]{chen2018attacking}
Hongge Chen, Huan Zhang, Pin-Yu Chen, Jinfeng Yi, and Cho-Jui Hsieh. 2018.
\newblock {Attacking visual
  language grounding with adversarial examples: A case study on neural image
  captioning}.
\newblock In \emph{Proceedings of the 56th Annual Meeting of the Association
  for Computational Linguistics (Volume 1: Long Papers)}, pages 2587--2597,
  Melbourne, Australia. Association for Computational Linguistics.

\bibitem[{Chen et~al.(2017)Chen, Zhu, Ling, Wei, Jiang, and
  Inkpen}]{ChenZLWJI17}
Qian Chen, Xiaodan Zhu, Zhen-Hua Ling, Si~Wei, Hui Jiang, and Diana Inkpen.
  2017.
\newblock  {Enhanced {LSTM} for
  natural language inference}.
\newblock In \emph{Proceedings of the 55th Annual Meeting of the Association
  for Computational Linguistics (Volume 1: Long Papers)}, pages 1657--1668,
  Vancouver, Canada. Association for Computational Linguistics.

\bibitem[{Conneau et~al.(2018)Conneau, Rinott, Lample, Williams, Bowman,
  Schwenk, and Stoyanov}]{conneau-etal-2018-xnli}
Alexis Conneau, Ruty Rinott, Guillaume Lample, Adina Williams, Samuel Bowman,
  Holger Schwenk, and Veselin Stoyanov. 2018.
\newblock {{XNLI}: Evaluating
  cross-lingual sentence representations}.
\newblock In \emph{Proceedings of the 2018 Conference on Empirical Methods in
  Natural Language Processing}, pages 2475--2485, Brussels, Belgium.
  Association for Computational Linguistics.

\bibitem[{Creutz(2018)}]{creutz-2018-open}
Mathias Creutz. 2018.
\newblock {Open subtitles
  paraphrase corpus for six languages}.
\newblock In \emph{Proceedings of the Eleventh International Conference on
  Language Resources and Evaluation ({LREC}-2018)}, Miyazaki, Japan. European
  Languages Resources Association (ELRA).

\bibitem[{Devlin et~al.(2019)Devlin, Chang, Lee, and
  Toutanova}]{devlin2018bert}
Jacob Devlin, Ming-Wei Chang, Kenton Lee, and Kristina Toutanova. 2019.
\newblock {{BERT}: Pre-training of
  deep bidirectional transformers for language understanding}.
\newblock In \emph{Proceedings of the 2019 Conference of the North {A}merican
  Chapter of the Association for Computational Linguistics: Human Language
  Technologies, Volume 1 (Long and Short Papers)}, pages 4171--4186,
  Minneapolis, Minnesota. Association for Computational Linguistics.

\bibitem[{Elliott et~al.(2016)Elliott, Frank, Sima{'}an, and
  Specia}]{desmond2016multi30k}
Desmond Elliott, Stella Frank, Khalil Sima{'}an, and Lucia Specia. 2016.
\newblock {{M}ulti30{K}:
  Multilingual {E}nglish-{G}erman image descriptions}.
\newblock In \emph{Proceedings of the 5th Workshop on Vision and Language},
  pages 70--74, Berlin, Germany. Association for Computational Linguistics.

\bibitem[{Glockner et~al.(2018)Glockner, Shwartz, and
  Goldberg}]{glockner2018breaking}
Max Glockner, Vered Shwartz, and Yoav Goldberg. 2018.
\newblock {Breaking {NLI} systems
  with sentences that require simple lexical inferences}.
\newblock In \emph{Proceedings of the 56th Annual Meeting of the Association
  for Computational Linguistics (Volume 2: Short Papers)}, pages 650--655,
  Melbourne, Australia. Association for Computational Linguistics.

\bibitem[{Iyyer et~al.(2018)Iyyer, Wieting, Gimpel, and
  Zettlemoyer}]{IyyerWGZ18}
Mohit Iyyer, John Wieting, Kevin Gimpel, and Luke Zettlemoyer. 2018.
\newblock {Adversarial example
  generation with syntactically controlled paraphrase networks}.
\newblock In \emph{Proceedings of the 2018 Conference of the North {A}merican
  Chapter of the Association for Computational Linguistics: Human Language
  Technologies, Volume 1 (Long Papers)}, pages 1875--1885, New Orleans,
  Louisiana. Association for Computational Linguistics.

\bibitem[{Jia and Liang(2017)}]{JiaL17}
Robin Jia and Percy Liang. 2017.
\newblock {Adversarial examples
  for evaluating reading comprehension systems}.
\newblock In \emph{Proceedings of the 2017 Conference on Empirical Methods in
  Natural Language Processing}, pages 2021--2031, Copenhagen, Denmark.
  Association for Computational Linguistics.

\bibitem[{Ribeiro et~al.(2018)Ribeiro, Singh, and
  Guestrin}]{ribeiro2018semantically}
Marco~Tulio Ribeiro, Sameer Singh, and Carlos Guestrin. 2018.
\newblock {Semantically equivalent
  adversarial rules for debugging {NLP} models}.
\newblock In \emph{Proceedings of the 56th Annual Meeting of the Association
  for Computational Linguistics (Volume 1: Long Papers)}, pages 856--865,
  Melbourne, Australia. Association for Computational Linguistics.

\bibitem[{Zhang et~al.(2019)Zhang, Baldridge, and He}]{paws2019naacl}
Yuan Zhang, Jason Baldridge, and Luheng He. 2019.
\newblock {{PAWS}: Paraphrase
  adversaries from word scrambling}.
\newblock In \emph{Proceedings of the 2019 Conference of the North {A}merican
  Chapter of the Association for Computational Linguistics: Human Language
  Technologies, Volume 1 (Long and Short Papers)}, pages 1298--1308,
  Minneapolis, Minnesota. Association for Computational Linguistics.

\bibitem[{Zhao et~al.(2018)Zhao, Dua, and Singh}]{zhao2018generating}
Zhengli Zhao, Dheeru Dua, and Sameer Singh. 2018.
\newblock {Generating natural
  adversarial examples}.
\newblock In \emph{International Conference on Learning Representations}.

\end{thebibliography}
\bibliographystyle{acl_natbib}

\end{document}